\documentclass[10pt]{IEEEtran}
\IEEEoverridecommandlockouts
\usepackage{cite}
\usepackage{amsmath,amssymb,amsfonts}
\usepackage{graphicx}
\usepackage{textcomp}
\usepackage{amsmath}
\usepackage{booktabs} 
\usepackage{multirow}
\usepackage{hyperref}       
\usepackage{url}            
\usepackage{booktabs}       
\usepackage{amsfonts}       
\usepackage{nicefrac}       
\usepackage{microtype}      
\usepackage{xcolor}         
\usepackage{algorithm}
\usepackage{algpseudocode}
\usepackage{amsmath}
\usepackage{tikz}

\usepackage{orcidlink}
\usepackage{comment}
\usepackage{amssymb}
\usepackage{braket}
\usepackage{stfloats}
\usepackage{orcidlink}
\usepackage[skip=5pt,font=small]{caption}
\usepackage{xcolor}
\def\BibTeX{{\rm B\kern-.05em{\sc i\kern-.025em b}\kern-.08em
    T\kern-.1667em\lower.7ex\hbox{E}\kern-.125emX}}

    \usepackage[table]{xcolor}

\definecolor{headergray}{RGB}{235,238,242}
\definecolor{rowgray}{RGB}{248,249,251}
\definecolor{highlightrow}{RGB}{235,245,238}
\begin{document}

\title{Q-SpiRL: Quantum Spiking Reinforcement Learning for Adaptive Robot Navigation}

\author{\IEEEauthorblockN{Mohamed Khair Altrabulsi\orcidlink{0009-0008-8753-2611}\textsuperscript{1,2}, Nouhaila Innan\orcidlink{0000-0002-1014-3457}\textsuperscript{1,2}, Alberto Marchisio\orcidlink{0000-0002-0689-4776}\textsuperscript{1,2},\\ Muhammad Kashif\orcidlink{0000-0003-2023-6371}\textsuperscript{1,2}, and Muhammad Shafique\orcidlink{0000-0002-2607-8135}\textsuperscript{1,2}\\
 \IEEEauthorblockA{
 \textsuperscript{1}eBRAIN Lab, Division of Engineering, New York University Abu Dhabi (NYUAD), Abu Dhabi, UAE\\
 \textsuperscript{2}Center for Quantum and Topological Systems (CQTS), NYUAD Research Institute, NYUAD, Abu Dhabi, UAE\\
 \{mka7870, nouhaila.innan, alberto.marchisio, muhammadkashif, muhammad.shafique\}@nyu.edu\\
}}}
\maketitle
\thispagestyle{empty}
\pagestyle{empty}

\begin{abstract}
Adaptive robot navigation in dynamic environments requires policies that can reach the target reliably while producing efficient and stable trajectories. This paper presents \textbf{Q-SpiRL}, a quantum spiking reinforcement learning framework for obstacle-aware robot navigation. The framework develops and evaluates five agent families: tabular Q-learning, classical MLP, classical SNN, quantum-enhanced MLP (QMLP), and quantum-enhanced spiking neural network (QSNN). While all models are implemented under a unified training and evaluation pipeline, the QSNN is the central architecture of interest, as it combines spike-based temporal processing with variational quantum feature transformation. Experiments are conducted across three grid-world environments of increasing size, namely 20$\times$20, 30$\times$30, and 40$\times$40, with both static and dynamic obstacles. Performance is assessed using success rate, success-weighted path length, path length, and turn rate under deterministic inference. Results show that QSNN achieves the strongest overall trade-off between task completion, trajectory efficiency, and motion smoothness, reaching up to 99\% success rate while maintaining high path efficiency in the most challenging setting. Execution on IBM quantum hardware further demonstrates the feasibility of deploying the proposed hybrid policy under real-device conditions.
\end{abstract}


\section{Introduction}

Autonomous robots are increasingly expected to operate in environments that change while they move. In such settings, an agent must make sequential decisions under uncertainty, react to moving obstacles, avoid collisions, and still reach the target through efficient and stable trajectories. This requirement is particularly challenging for embedded and resource-constrained robotic platforms, where navigation policies must be reliable while also remaining computationally efficient and suitable for low-power deployment \cite{desouza2002vision,chen2025survey,nahavandi2025comprehensive}.

Reinforcement learning (RL) provides a natural framework for learning such decision-making policies through interaction with the environment \cite{sutton1998reinforcement,zhu2021deep}. However, conventional tabular Q-learning becomes difficult to scale as the state-action space grows \cite{watkins1992q,mason2019review}, while deep RL methods often require substantial training data and computational resources \cite{botvinick2019reinforcement}. These limitations motivate the development of policy architectures that can preserve the decision-making capability of RL while improving efficiency, stability, and scalability in dynamic navigation tasks.

Spiking neural networks (SNNs) offer a promising direction for efficient policy learning because they process information through sparse, event-driven spike activity \cite{ghosh2009spiking,bing2018survey,zhang2025systematic}. This property has made them widely studied for energy-efficient computation, particularly in neuromorphic systems. In robot navigation, deep spiking Q-networks (DSQN) have shown that spiking RL agents can achieve competitive path-planning performance with reduced computational overhead \cite{Chen2022DeepRL,Zanatta2024,KUMAR2025129916}. In parallel, quantum machine learning (QML) has introduced parameterized quantum circuits and variational quantum models as trainable components with rich representational capacity \cite{Biamonte2017,chang2025primer}. These developments motivate a central question: \textit{can variational quantum processing improve the policy quality of spiking RL agents for obstacle-aware navigation?}

To address this question, this work proposes \textbf{Q-SpiRL} (\textbf{Q}uantum \textbf{Spi}king \textbf{R}einforcement \textbf{L}earning), a quantum-enhanced spiking reinforcement learning framework for robot navigation. The framework includes five agent families: tabular Q-learning, classical MLP, classical SNN, quantum-enhanced MLP (QMLP), and quantum-enhanced spiking neural network (QSNN). Among these agents, the QSNN is the central architecture of interest, as it integrates a variational quantum circuit into a DSQN-style spiking pipeline. Instead of applying quantum processing directly to raw state features, the QSNN first converts the discrete navigation state into spike-based temporal representations. The resulting firing-rate features are then processed by a parameterized quantum layer before action-value estimation. This design combines spike-based temporal processing with quantum feature transformation within a unified RL policy.

The proposed framework is evaluated across three grid-world environments of increasing size, namely 20$\times$20, 30$\times$30, and 40$\times$40, containing both static and dynamic obstacles. To ensure fair inference across all agents, each trained policy is converted into an explicit Q-table over the discrete state space and evaluated using deterministic greedy action selection. The models are assessed using success rate, success-weighted path length, path length, and turn rate, which jointly capture task completion, trajectory efficiency, and motion smoothness.

The main contributions of this work are summarized as follows:
\begin{itemize}
    \item We introduce \textbf{Q-SpiRL}, a quantum-enhanced spiking reinforcement learning framework for obstacle-aware robot navigation that includes tabular, classical neural, spiking, and quantum-enhanced agent families under a unified pipeline.
    
    \item We design the \textbf{QSNN} as the central policy architecture of the framework, integrating a variational quantum circuit into a DSQN-style spiking pipeline to transform spike-derived firing-rate representations before action-value estimation.
    
    \item We implement and compare five agent families: tabular Q-learning, classical MLP, classical SNN, QMLP, and QSNN, enabling a controlled study of dense, spiking, and quantum-enhanced policy architectures.
    
    \item We establish a deterministic evaluation protocol based on explicit Q-table conversion, allowing all trained agents to be evaluated under the same greedy inference mechanism over the full discrete state space.
    
    \item We evaluate the proposed framework across three navigation environments with increasing scale and obstacle complexity using success rate, success-weighted path length, path length, and turn rate.
    
    \item We show that QSNN achieves the strongest overall trade-off between task completion, trajectory efficiency, and motion smoothness, and we further demonstrate its feasibility through execution on IBM quantum hardware.
\end{itemize}

The remainder of this paper is organized as follows. Section ~\ref{sec2} reviews background and related work. Section~\ref{sec3} presents the proposed methodology, including the navigation environment, learning architectures, hyperparameter search, Q-table conversion, and evaluation protocol. Section~\ref{sec4} reports the experimental results, trajectory-level analysis, and quantum hardware evaluation. Section~\ref{sec5} concludes the paper and outlines future research directions.

\section{Background and Related Work}\label{sec2}

\subsection{Reinforcement Learning and Spiking RL for Navigation}

Reinforcement learning (RL) provides a standard framework for sequential decision-making, where an agent learns a policy through interaction with an environment. In robot navigation, RL is widely used for goal-reaching and obstacle-avoidance tasks because it allows agents to learn action policies from reward feedback rather than relying only on hand-crafted rules. Classical Q-learning is effective in small discrete settings, but its tabular formulation scales poorly as the state-action space grows \cite{watkins1992q}. Deep RL addresses this limitation by using neural networks as function approximators, enabling learning in larger and more complex environments. However, DRL often requires substantial training data, careful tuning, and significant computational resources, which can limit its use in embedded or energy-constrained robotic platforms \cite{11343694}.

SNNs provide an efficient alternative to conventional neural policies by using sparse spike events and temporal dynamics, making them suitable for low-power inference on neuromorphic hardware \cite{rathi2023exploring}. In reinforcement learning, DSQN-style methods integrate spiking dynamics into value-based policy learning and have been explored for navigation tasks under computational constraints \cite{KUMAR2025129916,Yang2025}. However, these approaches remain fully classical and do not examine whether variational quantum feature transformations can improve the policy quality of spiking RL agents in obstacle-aware navigation.

\subsection{Quantum Learning for Navigation and Decision-Making}

QML uses parameterized quantum circuits and hybrid quantum-classical models for representation learning, optimization, and decision-making \cite{innan2025lep,innan2025qnn,innan2025next}. Within this direction, Quantum RL extends these ideas to sequential decision-making through quantum policies, quantum-inspired action selection, or hybrid value estimation \cite{JerbiQRL,dutta2024qadqn,dutta2025qas,meyer2022survey,sefrin2025quantum}. In navigation and robotics, early quantum-inspired RL methods used quantum measurement and amplitude-amplification ideas to improve action selection and learning robustness in mobile robot navigation \cite{dong2010robust}. More recent studies have explored hybrid quantum deep reinforcement learning for robotic navigation, showing that parameterized quantum circuits can learn navigation policies with fewer trainable parameters than some classical baselines, although training stability and scalability remain challenging \cite{hohenfeld2024quantum}. Quantum-supported DRL has also been investigated for collision-free navigation in self-driving scenarios through hybrid quantum-classical critics \cite{sinha2025nav}, while quantum-classical RL frameworks have been proposed for dynamic path planning by combining quantum-generated Q-tables with classical reinforcement learning and turn-cost estimation \cite{tomar2025quantum}.

Quantum methods have also been extended to broader navigation and robotic decision-making settings. Quantum-based multi-agent reinforcement learning (QMARL) has been proposed to improve cooperative navigation through quantum state-action quantization, Grover-based action selection, and quantum prioritized experience replay \cite{chen2024qmarl}. In UAV navigation, QAOA-based path-planning frameworks such as QUAV formulate trajectory optimization as a quantum optimization problem and validate feasible path generation under simulation and quantum hardware execution \cite{innan2025quav}. Other related directions include quantum or hybrid optimization methods for mobile robot routing, such as quantum-enhanced ant colony optimization \cite{sarkar2024novel}, as well as broader studies on quantum robotics and quantum navigation systems \cite{yan2024quantum,sambataro2025current}.

Beyond navigation-specific studies, quantum spiking neural networks and quantum-enhanced spiking models have also been investigated as attempts to combine spike-based temporal processing with quantum representations or variational quantum circuits \cite{Pasquali2023,Brand2024,Khatoniar2024,innan2025fl,Liu2025,innan2026spate}. However, existing quantum navigation studies mostly focus on dense hybrid quantum-classical policies, quantum-inspired action selection, actor-critic learning, multi-agent coordination, or optimization-based path planning. They do not directly integrate variational quantum circuits with spiking RL architectures. This leaves a clear gap for studying whether spike-derived temporal representations can serve as an effective interface for quantum-enhanced policy learning in obstacle-aware navigation.

\subsection{Positioning of This Work}

This work addresses this gap by introducing \textbf{Q-SpiRL}, a quantum-enhanced spiking reinforcement learning framework for robot navigation. The central model, QSNN, combines spike-based temporal processing with a variational quantum layer for action-value estimation. Unlike prior quantum RL navigation studies that mainly rely on dense hybrid quantum-classical models, QSNN applies quantum processing to firing-rate representations derived from spiking dynamics. This allows us to study whether spike-based temporal representations provide an effective interface for quantum-enhanced policy learning.

The proposed study also differs from prior work in its evaluation design. Rather than comparing one quantum model against a single classical baseline, we implement and evaluate five agent families: tabular Q-learning, classical MLP, classical SNN, QMLP, and QSNN. All agents are tested under the same grid-world navigation formulation with static and dynamic obstacles and are evaluated through deterministic Q-table inference. This controlled setup allows a fair comparison across tabular, dense neural, spiking, quantum-dense, and quantum-spiking policies, while clarifying the role of the quantum layer in navigation performance.
\section{Our Q-SpiRL Methodology}\label{sec3}

\begin{figure*}[htpbt]
    \centering
    \includegraphics[width=\linewidth]{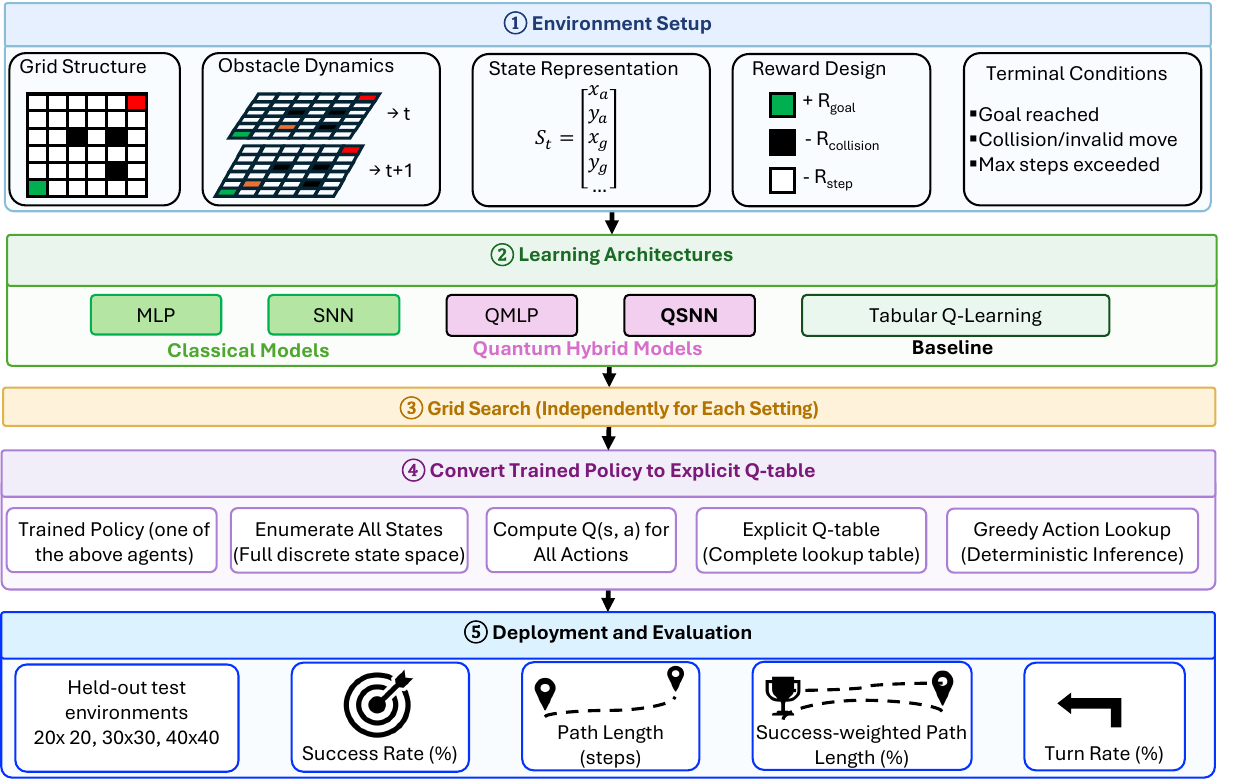}
\caption{Overview of the proposed methodology pipeline. The framework begins with the definition of the navigation environment. It then considers multiple learning architectures, with the QSNN as the central quantum-enhanced model of this work. After independent hyperparameter selection through grid search, each trained policy is converted into an explicit Q-table for deterministic greedy inference. Final evaluation is carried out on held-out environments of different sizes using success rate, path length, success-weighted path length, and turn rate.}
    \label{methodology}
\end{figure*}

The proposed methodology follows a multi-stage pipeline designed to enable a fair and controlled comparison between classical and quantum-enhanced reinforcement learning agents for grid-based navigation, as illustrated in Fig. \ref{methodology}. First, a discrete navigation environment is defined, including the grid structure, obstacle dynamics, state representation, reward design, and terminal conditions. This establishes a common decision-making framework for all considered agents. Second, four neural policy families are developed: a classical MLP, a classical SNN, a QMLP, and a QSNN, alongside a tabular Q-learning baseline serving as a non-neural reference. 

Among these models, the QSNN constitutes the central contribution of this work and serves as the primary architecture of interest in the proposed quantum-enhanced framework. Third, model-specific and architecture-specific hyperparameters are selected through grid search independently for each environment size and for each classical or quantum setting. Fourth, after training, each learned policy is converted into an explicit Q-table over the full discrete state space, enabling deterministic inference through greedy action lookup and eliminating stochastic effects during deployment. Finally, all methods are evaluated on held-out environments using success rate, path length, success-weighted path length, and turn rate. This organization ensures that performance differences can be attributed to the learning architecture itself, particularly to the inclusion of quantum variational processing, rather than to inconsistencies in inference or evaluation protocols.

\subsection{Navigation Environment and RL Formulation}

The navigation task is formulated as a discrete reinforcement learning problem in a two-dimensional grid world inspired by the DSQN environment design proposed by Kumar \emph{et al.} \cite{KUMAR2025129916}. The same environment definition, state formulation, reward structure, and terminal conditions are used for all classical and quantum-enhanced agents to ensure a fair comparison across model families. In each episode, the agent must navigate from a fixed start cell to a fixed target cell while avoiding both static and dynamic obstacles under a discrete action space.

The environment is defined on a finite grid
\begin{equation}
\mathcal{G} = \{0,\dots,g-1\} \times \{0,\dots,g-1\},
\end{equation}
where $g$ denotes the grid size. In this work, we consider $g \in \{20,30,40\}$, corresponding to the $20\times20$, $30\times30$, and $40\times40$ environments, respectively. At time step $t$, the agent is characterized by its current position $c_t \in \mathcal{G}$ and heading angle $\theta_t \in [-\pi,\pi]$. The agent starts from a fixed initial cell $c_0$ and aims to reach a fixed terminal goal cell $c_T$.

The action space consists of five discrete actions
\begin{equation}
\mathcal{A} = \{a_1,a_2,a_3,a_4,a_5\}, \qquad |\mathcal{A}| = 5,
\end{equation}
which correspond to relative heading changes
\begin{equation}
\Delta \theta_a \in \left\{-\frac{\pi}{2}, -\frac{\pi}{4}, 0, \frac{\pi}{4}, \frac{\pi}{2}\right\}.
\end{equation}
After selecting action $a_t$, the heading is updated as
\begin{equation}
\theta_{t+1} = \theta_t + \Delta \theta_{a_t},
\end{equation}
and the agent moves by one grid cell in the resulting direction. This formulation allows both axis-aligned and diagonal motion while preserving a compact and interpretable action space. Each episode is truncated after a maximum of $T_{\max}=300$ steps if no terminal condition is reached.

The environment contains both static obstacles $\mathcal{O}_s$ and dynamic obstacles $\mathcal{O}_d$, which are sampled at the beginning of each episode while excluding the start and goal cells. Static obstacles remain fixed throughout the episode, whereas dynamic obstacles evolve according to
\begin{equation}
o^{(d)}_{t+1} = o^{(d)}_t + \delta_d,
\end{equation}
where
\begin{equation}
\delta_d \in \{(1,0), (-1,0), (0,1), (0,-1)\}.
\end{equation}
Thus, dynamic obstacles move along the four cardinal directions and are updated at half the speed of the agent, that is, once every two agent steps. Whenever a boundary crossing or a conflicting move would occur, the obstacle reverses its direction. To avoid trivial shortest-path solutions, the obstacle placement strategy is designed to frequently obstruct the direct diagonal route between the start and target locations, thereby enforcing non-trivial obstacle-avoidance behavior.

Three environment scales are considered in this work: $20\times20$, $30\times30$, and $40\times40$. For the $20\times20$ environment, we use 6 randomized static obstacles, one additional training-only diagonal static obstacle, and 1 randomized dynamic obstacle. For the $30\times30$ environment, we use 9 randomized static obstacles, one additional training-only diagonal static obstacle, and 1 randomized dynamic obstacle. For the $40\times40$ environment, we use 12 randomized static obstacles, one additional training-only diagonal static obstacle, and 3 randomized dynamic obstacles. These progressively larger settings increase both the reachable space and the obstacle complexity, allowing the learned policies to be evaluated under more challenging navigation conditions.

Rather than observing the full grid directly, the agent receives a compact discrete state vector
\begin{equation}
s_t = (R_o, D_o, R_T, A_{T\rightarrow o}), \label{eqstate}
\end{equation}
where $R_o \in \{1,\dots,8\}$ denotes the angular region of the nearest obstacle relative to the current heading of the agent, while $R_T \in \{1,\dots,8\}$ denotes the angular region of the target. The component $A_{T\rightarrow o} \in \{1,\dots,8\}$ is a discretized angular bin of the angle
\begin{equation}
\alpha_t =\arccos\left(\frac{(c_T-c_t)^\top (o_t-c_t)}{\|c_T-c_t\|_2 \, \|o_t-c_t\|_2}\right),
\end{equation}
which captures the geometric relation between the target direction and the obstacle direction as perceived from the current agent position. Finally, $D_o \in \{0,1,2,3,4\}$ encodes the motion direction of the nearest dynamic obstacle, where $D_o = 0$ indicates either a static obstacle or the absence of a nearby dynamic obstacle. 

The reward function is designed to encourage progress toward the target while discouraging unsafe behavior near obstacles. At time step $t$, the reward is defined as
\begin{equation}
r_t=\beta_1(d_{t-1}-d_t)+\exp\bigl(\beta_2(d^o_{t-1}-d^o_t)\bigr)+\beta_3 \sin(\alpha_t),
\end{equation}
where $d_t = \|c_T-c_t\|_2$ denotes the Euclidean distance to the target, $d^o_t$ denotes the distance to the nearest obstacle, and $\beta_2<0$ controls the obstacle-avoidance penalty. The first term rewards progress toward the goal by favoring reductions in target distance. Since $\beta_2<0$, moving closer to an obstacle makes $(d^o_{t-1}-d^o_t)>0$, which reduces the exponential term and therefore discourages unsafe motion near obstacles. The third term encourages angular separation between the target direction and the obstacle direction, helping the agent avoid moving directly toward obstructed regions. In addition, a small per-step penalty is applied to favor shorter and more efficient trajectories.

An episode terminates when the agent reaches the target $c_t = c_T,$
or when it collides with a static or dynamic obstacle, leaves the grid boundary, or exceeds the maximum episode length. Reaching the goal yields a large positive terminal reward, whereas collisions and boundary violations incur a large negative terminal reward. Episodes that terminate due to timeout are also penalized to discourage indecisive or excessively long trajectories.

\subsection{Learning Architectures}

To assess the effect of quantum variational processing in grid-based reinforcement learning, we consider five policy families: a classical multilayer perceptron (MLP), a SNN, a quantum-enhanced multilayer perceptron (QMLP), QSNN, and a tabular Q-learning baseline. All neural agents operate on the same discrete state formulation and produce action-value estimates over the same five-action decision space. To ensure a fair comparison, the classical MLP and SNN are parameter-matched, while the QMLP and QSNN are likewise designed with comparable capacity. Among the considered neural models, the QSNN is the central contribution of this work and represents the primary quantum-enhanced architecture under study.

\subsubsection{Classical MLP}
\begin{figure}[htbp]
    \centering
    \includegraphics[width=\columnwidth]{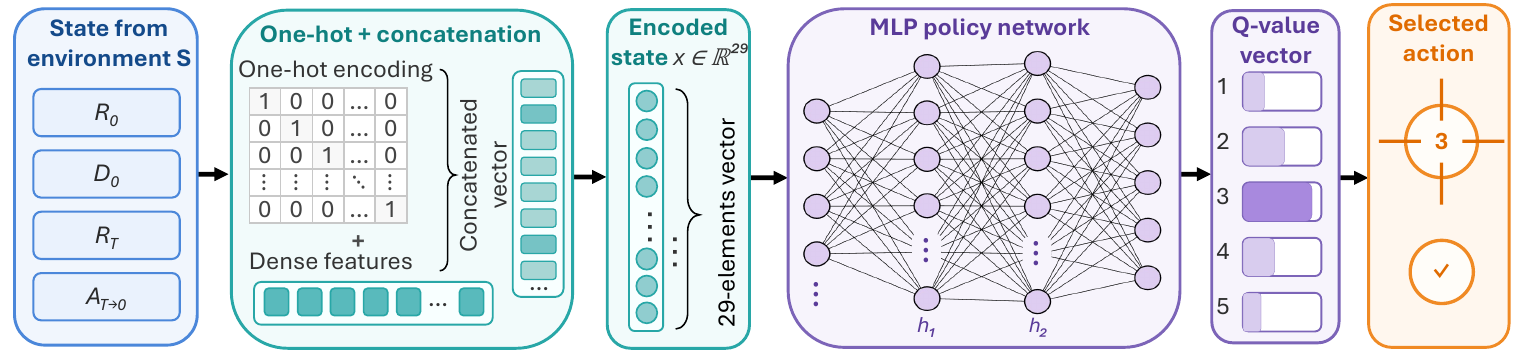}
    \caption{Architecture of the classical MLP.}
    \label{fig:mlp_diagram}
\end{figure}

The classical multilayer perceptron serves as a standard non-spiking neural baseline. The discrete environment state $s \in \mathbb{R}^{4}$ is mapped to an input vector $x \in \mathbb{R}^{29}$ through one-hot encoding and concatenation, reflecting the finite and discrete structure of each state component. This encoded representation is processed by a stack of fully connected layers with nonlinear activations, followed by a final linear output layer that produces five Q-values, one for each action (see Fig.~\ref{fig:mlp_diagram}). The model uses 30 hidden neurons and contains 1,985 trainable parameters, matching the capacity of the classical SNN.
\subsubsection{Classical SNN}
\begin{figure}[htbp]
    \centering
    \includegraphics[width=\columnwidth]{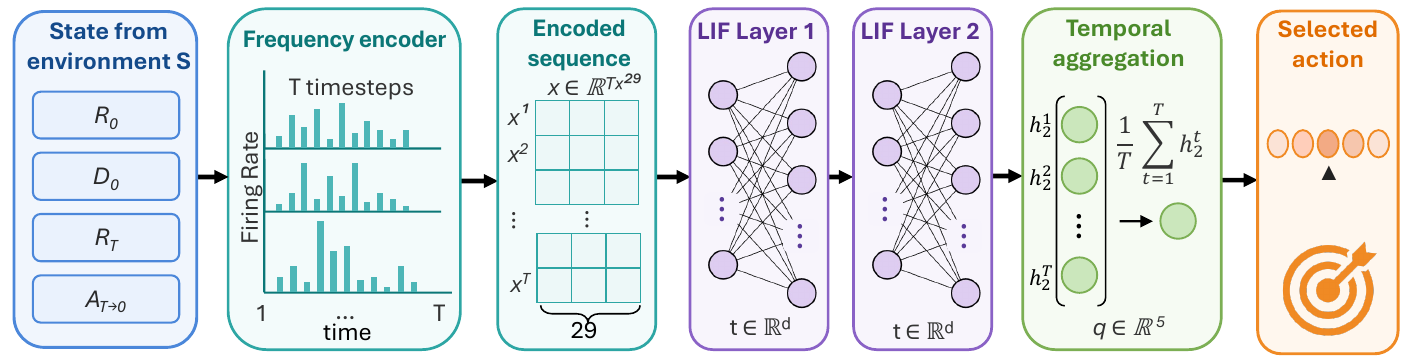}
    \caption{Architecture of the classical SNN.}
    \label{fig:snn_diagram}
\end{figure}

The classical SNN replaces conventional pointwise nonlinearities with leaky integrate-and-fire (LIF) neurons and processes the encoded state through temporal spike dynamics. As in the MLP case, the discrete state is first transformed into the same 29-dimensional one-hot representation and is then converted into a spike-based input sequence using a frequency-based Poisson encoder. For each neuron, the spike probability is defined as
\begin{equation}
p_j = 1 - \exp(-r_j \Delta t),
\end{equation}
where $r_j$ denotes the assigned firing rate and $\Delta t$ is the simulation step size. The resulting spike train is processed by stacked linear layers followed by LIF cells with membrane and synaptic dynamics (see Fig.~\ref{fig:snn_diagram}). The outputs of the final spiking layer are aggregated across time using mean pooling,
\begin{equation}
\bar{o} = \frac{1}{T} \sum_{t=1}^{T} o^t,
\end{equation}
and a final non-spiking linear layer maps the aggregated response to five Q-values. The network uses 30 hidden neurons and contains 1,985 trainable parameters, matching the classical MLP.
\subsubsection{Quantum-Enhanced MLP (QMLP)}
\begin{figure}[htbp]
    \centering
    \includegraphics[width=1\linewidth]{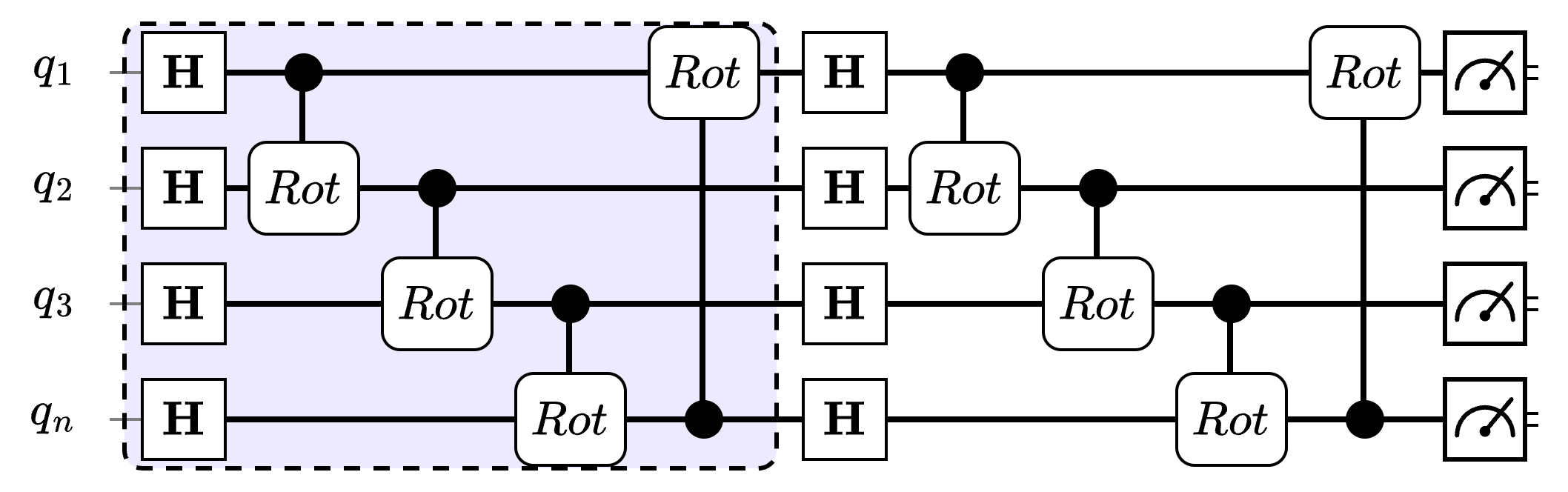}
    \caption{Variational quantum circuit, also referred to as a parameterized quantum circuit, used in both the QMLP and QSNN architectures. The circuit is composed of repeated variational blocks acting on $n$ qubits, where the highlighted region represents a single layer $L$. Each layer consists of Hadamard gates, trainable controlled rotation gates, and entangling operations, and the full circuit is formed by stacking multiple such layers. Final measurements on all qubits produce the quantum features used by the hybrid model.}
    \label{fig:vqc}
\end{figure}

The quantum-enhanced multilayer perceptron extends the classical MLP by inserting a variational quantum circuit between classical preprocessing and output layers. The one-hot encoded input vector $x \in \mathbb{R}^{29}$
is first projected by classical fully connected layers into a latent feature vector whose dimension matches the number of qubits, with $q = 8.$
This latent representation is then encoded into a parameterized quantum circuit composed of variational layers (see Fig. \ref{fig:vqc}). The circuit produces quantum features through Pauli-$Z$ expectation values,
\begin{equation}
z_q = \langle 0 | U(x;\Theta)^\dagger \left(I^{\otimes(q-1)} \otimes Z_q \right) U(x;\Theta) | 0 \rangle.
\end{equation}
These quantum features are subsequently processed by post-quantum classical layers to produce the five output Q-values. The QMLP contains 1,977 trainable parameters and is parameter-matched with the QSNN.
\subsubsection{Quantum-Enhanced SNN (QSNN)}
The quantum-enhanced SNN is the main model proposed in this work. It integrates a variational quantum circuit into a spiking RL pipeline, combining temporal spike-based processing with quantum feature transformation. As in the classical SNN, the one-hot encoded state is first converted into a spike train through a frequency-based Poisson encoder and processed by pre-quantum spiking layers. The outputs of the final pre-quantum spiking layer are then aggregated across time to obtain a continuous firing-rate representation,
\begin{equation}
\bar{o}_{\mathrm{preQ}} = \frac{1}{T} \sum_{t=1}^{T} o^t_{\mathrm{preQ}} \in [0,1]^8.
\end{equation}
This rate vector is passed to a parameterized quantum circuit with 8 qubits, which generates quantum features through Pauli-$Z$ expectation values (see Fig. \ref{fig:vqc}). These quantum outputs are then mapped to five Q-values by post-quantum classical layers. By embedding a variational quantum module within the spiking decision pipeline, the QSNN is designed to capture both temporal structure and quantum-transformed latent representations within a unified policy architecture. The model uses 35 hidden neurons and contains 1,977 trainable parameters, matching the QMLP.

\subsubsection{Tabular Q-Learning Baseline}
As a non-neural reference, we include a tabular Q-learning agent that explicitly stores action values for every discrete state-action pair using the compact state representation defined in Eq.~\eqref{eqstate}. This gives a total discrete state space size of
\begin{equation}
|\mathcal{S}| = 8 \times 5 \times 8 \times 8 = 2560.
\end{equation}
The corresponding Q-table therefore has dimensions $\mathbb{R}^{|\mathcal{S}| \times |\mathcal{A}|},$ where $|\mathcal{A}| = 5$. During training, actions are selected using an $\epsilon$-greedy policy and Q-values are updated through standard temporal-difference learning. During evaluation, the learned Q-table induces a deterministic policy by selecting the greedy action at each state.

\subsection{Hyperparameter Search and Training Procedure}

To obtain competitive and fair model configurations, we perform a structured grid search over the main architectural and temporal hyperparameters of the considered agents. The search is conducted independently for the classical and quantum settings and separately for each environment scale, namely $20\times20$, $30\times30$, and $40\times40$. This design avoids transferring hyperparameter choices across substantially different navigation conditions and ensures that each model family is evaluated under its own best-performing configuration.

For all neural agents, training is carried out for 800 episodes within the corresponding environment setting. Candidate configurations are first assessed in terms of task completion reliability. Only models achieving a success rate of at least $95\%$ are retained for further comparison. Among the retained configurations, the final model selection is based on the mean path length over successful episodes, thereby favoring policies that are not only reliable but also efficient in their navigation behavior.

The search space includes both general neural-network choices and spiking-specific temporal parameters. For the MLP-based models, we consider the activation function as a tunable design choice. For the spiking models, the search includes the maximum firing rate, the number of simulation timesteps, the simulation step size, and the membrane and synaptic time constants. The resulting search space and the selected values for each environment and model family are summarized in Table~\ref{tab:hyperparams}.

\begin{table*}[htpbt]
\centering
\caption{Grid-search space and selected hyperparameter values for the classical and quantum agents across the three navigation environments.}
\label{tab:hyperparams}
\begin{tabular}{llc|ccc|ccc}
\toprule
 &  & \textbf{Search Values} 
 & \multicolumn{3}{c|}{\textbf{Classical }} 
 & \multicolumn{3}{c}{\textbf{Quantum }} \\
\cmidrule(lr){4-6} \cmidrule(lr){7-9}
\textbf{Group} & \textbf{Hyperparameter} 
& 
& 20$\times$20 & 30$\times$30 & 40$\times$40
& 20$\times$20 & 30$\times$30 & 40$\times$40 \\
\midrule

\multirow{5}{*}{\textbf{SNN}}
& $f_{\max}$ (Hz) & $\{20,100,200\}$ 
& 100 & 100 & 100 
& 100 & 100 & 200 \\

& $T$ & $\{5,10,15,20,30\}$ 
& 10 & 5 & 20 
& 10 & 20 & 5 \\

& $\Delta t$ & $\{0.01,0.05,0.1,0.2\}$ 
& 0.20 & 0.05 & 0.20 
& 0.10 & 0.05 & 0.01 \\

& $\tau_{\mathrm{mem}}$ & $\{0.01,0.02,0.04\}$ 
& 0.02 & 0.01 & 0.04 
& 0.04 & 0.02 & 0.01 \\

& $\tau_{\mathrm{syn}}$ & $\{0.005,0.01,0.02\}$ 
& 0.01 & 0.005 & 0.02 
& 0.01 & 0.005 & 0.005 \\

\midrule

\multirow{1}{*}{\textbf{MLP}}
& Activation & $\{\text{ReLU},\text{Tanh},\text{Sigmoid}\}$ 
& ReLU & ReLU & ReLU 
& ReLU & ReLU & ReLU \\

\bottomrule
\end{tabular}
\end{table*}

Classical and quantum variants are tuned separately rather than being forced to share identical hyperparameter settings, while parameter-matched architecture pairs are maintained at the model-design level. As a result, performance differences are less likely to arise from under-tuned configurations and can be interpreted more directly in relation to the underlying learning architecture.

\subsection{Inference Q-Table Construction}

To enable fair and fully deterministic deployment-time inference across all considered agents, each trained policy is converted into an explicit Q-table over the full discrete state space. This step provides a unified action-selection mechanism for all model families, regardless of whether the original policy is represented by a classical neural network, a spiking neural network, a quantum-enhanced model, or a tabular agent. In addition to improving comparability, this conversion also reduces inference cost at deployment time by replacing repeated forward passes with direct table lookup.

The Q-table is defined over the compact state representation in Eq.~\eqref{eqstate}.
Since $|\mathcal{S}| = 2560$ and $|\mathcal{A}| = 5$, the constructed table stores one action-value vector for each discrete state,
\begin{equation}
Q_{\mathrm{table}} \in \mathbb{R}^{|\mathcal{S}| \times |\mathcal{A}|}.
\end{equation}

For each trained model, the Q-table is populated by exhaustively enumerating all discrete states and querying the corresponding policy for its predicted action values. For MLP-based agents, namely the classical MLP and the QMLP, each state is deterministically mapped to its one-hot encoded input vector $x \in \mathbb{R}^{29}$, and the table entries are assigned as
\begin{equation}
Q_{\mathrm{table}}(s,\cdot) \leftarrow Q_{\mathrm{model}}(x).
\end{equation}

For spiking-based agents, namely the classical SNN and the QSNN, each discrete state is first converted into a spike-based input sequence using the same frequency-based Poisson encoding employed during training. The resulting spike tensor $X$ is then passed through the trained model, and the corresponding Q-values are stored as
\begin{equation}
Q_{\mathrm{table}}(s,\cdot) \leftarrow Q_{\mathrm{model}}(X).
\end{equation}
A single offline encoding realization is used during this conversion step. As a result, stochasticity associated with spike generation is confined to table construction and does not affect deployment-time inference.

Once the table has been constructed, a deterministic greedy policy is obtained as
\begin{equation}
\pi(s) = \arg\max_{a \in \mathcal{A}} Q_{\mathrm{table}}(s,a).
\end{equation}
Thus, during evaluation, action selection is performed exclusively through direct lookup, which ensures deterministic deployment-time inference and directly comparable evaluation across all methods.
\subsection{Test Protocol}

All methods are evaluated on a held-out set of test environments generated by varying the environment seed while keeping the environment definition fixed. Specifically, we consider $N_{\mathrm{env}} = 100$ test environments associated with the seed set
\begin{equation}
\mathcal{Z}_{\mathrm{test}} = \{2000, 2001, \dots, 2099\}.
\end{equation}
This protocol ensures that all agents are tested on the same collection of unseen obstacle configurations under identical task conditions.

For each test environment, a single episode is executed with a maximum horizon of $T_{\max} = 300$ steps. At the beginning of each episode, the environment is reset using the corresponding test seed, and the agent acts according to the deterministic greedy policy obtained from the converted Q-table. Thus, action selection during testing is performed as
\begin{equation}
a_t = \pi(s_t),
\end{equation}
with no stochastic exploration or sampling-based inference.

Each episode terminates upon reaching the target, colliding with a static or dynamic obstacle, leaving the grid boundary, or exceeding the maximum episode length. By evaluating all agents under the same deterministic inference protocol and the same held-out environments, the proposed test procedure enables a direct and fair comparison of navigation reliability, path efficiency, and behavioral stability across classical, spiking, and quantum-enhanced models.

\subsection{Evaluation Metrics}

We evaluate each learned policy using four metrics that capture task completion, path efficiency, and motion regularity. All metrics are computed over the held-out test environments under deterministic greedy inference derived from the constructed Q-table. These measures provide a balanced assessment of whether an agent can reliably reach the target, how efficiently it moves, and how stable its action sequence remains during navigation.

For an executed trajectory
\begin{equation}
\rho_i = (p_{i,0}, p_{i,1}, \dots, p_{i,T_i})
\end{equation}
in episode $i$, the path length is defined as
\begin{equation}
L_i = \sum_{t=1}^{T_i} c(p_{i,t-1}, p_{i,t}),
\end{equation}
where the transition cost depends on the type of movement,
\begin{equation}
c(p_{t-1}, p_t) =
\begin{cases}
\Delta, & \text{for axis-aligned motion}, \\
\sqrt{2}\,\Delta, & \text{for diagonal motion}.
\end{cases}
\end{equation}
Here, $\Delta$ denotes the side length of a grid cell. Unless otherwise stated, we report the mean path length over successful episodes,
\begin{equation}
\bar{L}_{\mathrm{succ}} =
\frac{1}{\sum_{i=1}^{N} S_i}
\sum_{i=1}^{N} S_i L_i,
\end{equation}
where $S_i \in \{0,1\}$ is the success indicator for episode $i$.

The success rate measures the fraction of evaluation episodes in which the agent reaches the target within the maximum horizon and without collision or boundary violation. Formally, it is defined as
\begin{equation}
\mathrm{SR} = \frac{1}{N} \sum_{i=1}^{N} S_i.
\end{equation}
This metric captures the reliability of task completion across unseen environments.

To jointly evaluate success and trajectory efficiency, we use Success-weighted Path Length (SPL). Let $L_i$ denote the executed path length, let $L_i^\star$ denote the shortest feasible path length for the same environment, and let $S_i$ denote the success indicator. SPL is defined as
\begin{equation}
\mathrm{SPL} =
\frac{1}{N}
\sum_{i=1}^{N}
S_i \frac{L_i^\star}{\max(L_i, L_i^\star)}.
\end{equation}
The SPL score ranges from 0 to 1 and rewards policies that both succeed and follow near-optimal paths. Failed episodes contribute zero.

To quantify motion smoothness and behavioral stability, we measure the turn rate based on the executed action sequence. Let $a_{i,t}$ denote the action selected at step $t$ in episode $i$. We define the turn indicator as
\begin{equation}
T(a_{i,t}) = \mathbb{1}[a_{i,t} \neq a_{\mathrm{forward}}],
\end{equation}
where $a_{\mathrm{forward}}$ denotes the zero-yaw action. The per-episode turn rate is then
\begin{equation}
\mathrm{TR}_i =
\frac{1}{T_i}
\sum_{t=1}^{T_i} T(a_{i,t}),
\end{equation}
and the reported turn rate is the mean over successful episodes,
\begin{equation}
\mathrm{TR} =
\frac{1}{\sum_{i=1}^{N} S_i}
\sum_{i=1}^{N} S_i \mathrm{TR}_i.
\end{equation}
Lower turn rate indicates smoother and less oscillatory navigation behavior.

\section{Results and Discussion}\label{sec4}

\subsection{Experimental Setup}

All models were trained on a high-performance computing (HPC) cluster using SLURM for job scheduling. Training was performed using CPU-only resources, without GPU acceleration. After training, each learned policy was converted into its explicit Q-table representation, and final evaluation was carried out locally on an Apple M4 MacBook.

Table~\ref{tab:fixed_hyperparams} summarizes the fixed hyperparameters used across experiments, including architectural settings, optimization parameters, replay and exploration settings, and environment-related constants. These values were kept unchanged across the evaluated models unless otherwise stated. Hyperparameters selected through the model-specific grid search, such as the spiking and temporal parameters of the SNN-based agents, are reported separately in the methodology section and are therefore not repeated here.

For the quantum-enhanced models, the variational circuit was configured with $q=8$ qubits and $L=3$ variational layers. The classical and quantum neural architectures used 30 and 35 hidden neurons, respectively, in order to maintain comparable capacity within each model family. All neural agents were trained for 800 episodes, and evaluation was performed on 100 held-out environments using deterministic greedy inference derived from the converted Q-tables.

\begin{table}[htpbt]
\centering
\caption{Fixed hyperparameters used across training and evaluation.}

\label{tab:fixed_hyperparams}
\begin{tabular}{p{1.5cm}ll}
\toprule
\textbf{Group} & \textbf{Hyperparameter} & \textbf{Value} \\
\midrule
\multirow{2}{*}{\textbf{Architecture}}
& Hidden neurons (Classical)  & 30 \\
& Hidden neurons (Quantum)    & 35 \\
\midrule
\multirow{2}{*}{\begin{tabular}{@{}c@{}}\textbf{Quantum}\\\textbf{Circuit}\end{tabular}}
& Qubits $q$                  & 8 \\
& Variational layers $L$      & 3 \\
\midrule
\multirow{4}{*}{\textbf{Exploration}}
& $\varepsilon$ (random valid action prob.) & 0.01 \\
& Initial temperature $T_0$   & 1.0 \\
& Temperature decay $\lambda_T$ & 0.999 \\
& Minimum temperature $T_{\min}$ & 0.05 \\
\midrule
\multirow{3}{*}{\textbf{Optimization}}
& Learning rate               & 0.005 \\
& Optimizer                   & Adam \\
& Gradient clip norm          & 1.0 \\
\midrule
\multirow{4}{*}{\textbf{DQN}}
& Discount factor $\gamma$    & 0.9 \\
& Replay buffer capacity      & $10^5$ \\
& Batch size                  & 128 \\
& Target network update freq. & 500 steps \\
\midrule
\multirow{2}{*}{\textbf{Training}}
& Episodes                    & 800 \\
& Learning start (steps)      & 1000 \\
\midrule
\multirow{2}{*}{\textbf{Environment}}
& Max steps per episode $T_{\max}$ & 300 \\
& Dynamic obstacle update freq. & Every 2 agent steps \\
\bottomrule
\end{tabular}
\end{table}
\subsection{Cross-Environment Performance}

We first examine the aggregate performance of all evaluated agents across the three navigation environments in order to identify the dominant trends before moving to the environment-specific analyses. The comparison focuses on four metrics that jointly capture task completion, trajectory efficiency, and motion regularity, namely success rate (SR), success-weighted path length (SPL), path length (PL), and turn rate (TR).

Table~\ref{tab:main_results} reports the quantitative results for all methods under the unified deterministic evaluation protocol. Across the evaluated environment scales, the proposed QSNN provides the strongest overall balance between reliability, path quality, and smoothness. More broadly, the spiking architectures remain more robust than the dense MLP-based models as the environment size increases, and the quantum-enhanced spiking model either matches or improves upon the classical SNN in all settings. By contrast, the Q-table baseline often attains competitive success rates but consistently yields longer and more oscillatory trajectories, which lowers its overall path quality. These observations indicate that success rate alone is insufficient to characterize navigation quality and that the QSNN provides the most favorable trade-off when all four metrics are considered jointly.
\begin{table*}[htpbt]
\centering
\caption{Performance comparison across the three navigation environments. Results are reported as mean $\pm$ standard error over 100 evaluation episodes. SR denotes success rate, SPL denotes success-weighted path length, PL denotes path length, and TR denotes turn rate. Higher is better for SR and SPL, while lower is better for PL and TR. Best values within each environment are shown in bold. Since PL and TR are computed over successful episodes only, they should be interpreted jointly with SR and SPL.}
\label{tab:main_results}
\begin{tabular}{
l
>{\centering\arraybackslash}p{2cm}
>{\centering\arraybackslash}p{2.5cm}
>{\centering\arraybackslash}p{2.5cm}
>{\centering\arraybackslash}p{2.5cm}
>{\centering\arraybackslash}p{2.5cm}
}
\toprule
\rowcolor{headergray}
\textbf{Environment} & \textbf{Model} & \textbf{SR $\uparrow$} & \textbf{SPL $\uparrow$} & \textbf{PL (m) $\downarrow$} & \textbf{TR $\downarrow$} \\
\midrule

\multirow{5}{*}{20$\times$20}
& Q-Table & 0.930 $\pm$ 0.0256 & 0.8068 $\pm$ 0.0243 & 18.5853 $\pm$ 0.2662 & 0.4857 $\pm$ 0.0160 \\
\rowcolor{rowgray}
& MLP     & 0.940 $\pm$ 0.0239 & 0.8519 $\pm$ 0.0227 & 17.6262 $\pm$ 0.1518 & 0.3247 $\pm$ 0.0201 \\
& SNN     & 0.950 $\pm$ 0.0219 & 0.8765 $\pm$ 0.0211 & 17.2842 $\pm$ 0.1253 & 0.2625 $\pm$ 0.0158 \\
\rowcolor{rowgray}
& QMLP    & 0.890 $\pm$ 0.0314 & 0.8596 $\pm$ 0.0307 & \textbf{16.4759 $\pm$ 0.1103} & 0.2670 $\pm$ 0.0192 \\
\rowcolor{highlightrow}
& QSNN    & \textbf{0.960 $\pm$ 0.0197} & \textbf{0.8933 $\pm$ 0.0192} & 17.1272 $\pm$ 0.1165 & \textbf{0.2282 $\pm$ 0.0135} \\
\midrule

\multirow{5}{*}{30$\times$30}
& Q-Table & \textbf{0.990 $\pm$ 0.0100} & 0.8271 $\pm$ 0.0119 & 29.8580 $\pm$ 0.3589 & 0.3821 $\pm$ 0.0124 \\
\rowcolor{rowgray}
& MLP     & 0.940 $\pm$ 0.0239 & 0.8463 $\pm$ 0.0224 & 27.5379 $\pm$ 0.2282 & 0.2853 $\pm$ 0.0159 \\
& SNN     & 0.980 $\pm$ 0.0141 & 0.8871 $\pm$ 0.0142 & 27.3731 $\pm$ 0.2019 & 0.2923 $\pm$ 0.0145 \\
\rowcolor{rowgray}
& QMLP    & 0.980 $\pm$ 0.0141 & 0.8760 $\pm$ 0.0142 & 27.7351 $\pm$ 0.2141 & 0.2657 $\pm$ 0.0132 \\
\rowcolor{highlightrow}
& QSNN    & 0.970 $\pm$ 0.0171 & \textbf{0.9166 $\pm$ 0.0166} & \textbf{26.1334 $\pm$ 0.1103} & \textbf{0.2512 $\pm$ 0.0109} \\
\midrule

\multirow{5}{*}{40$\times$40}
& Q-Table & \textbf{0.990 $\pm$ 0.0100} & 0.8477 $\pm$ 0.0097 & 39.1974 $\pm$ 0.2143 & 0.4766 $\pm$ 0.0100 \\
\rowcolor{rowgray}
& MLP     & 0.770 $\pm$ 0.0423 & 0.7600 $\pm$ 0.0418 & \textbf{33.8613 $\pm$ 0.1172} & \textbf{0.0607 $\pm$ 0.0082} \\
& SNN     & 0.980 $\pm$ 0.0141 & 0.8849 $\pm$ 0.0137 & 37.1888 $\pm$ 0.2217 & 0.3236 $\pm$ 0.0143 \\
\rowcolor{rowgray}
& QMLP    & 0.930 $\pm$ 0.0256 & 0.8236 $\pm$ 0.0233 & 37.9331 $\pm$ 0.2356 & 0.2973 $\pm$ 0.0134 \\
\rowcolor{highlightrow}
& QSNN    & \textbf{0.990 $\pm$ 0.0100} & \textbf{0.9023 $\pm$ 0.0105} & 36.8400 $\pm$ 0.2104 & 0.2604 $\pm$ 0.0129 \\
\bottomrule
\end{tabular}
\end{table*}
\subsubsection{20$\times$20 Environment}

In the 20$\times$20 environment, the QSNN achieves the strongest overall performance, leading simultaneously in success rate, SPL, and turn rate while remaining close to the shortest path length. Specifically, QSNN reaches a success rate of 96\%, the highest SPL at 0.893, and the lowest turn rate at 0.228 turns per step. By contrast, the Q-table baseline ranks below the neural approaches on all quality-related metrics, producing the lowest SPL (0.807), the longest paths (18.59\,m), and the highest turn rate (0.486), which indicates highly oscillatory behavior despite its competitive success rate of 93\%.

A direct comparison between QSNN and its classical SNN counterpart shows consistent gains across all four metrics. Relative to SNN, QSNN improves success rate (96\% vs.\ 95\%), SPL (0.893 vs.\ 0.876), path length (17.13\,m vs.\ 17.28\,m), and turn rate (0.228 vs.\ 0.263), indicating that the quantum layer provides a uniform benefit in the spiking setting without introducing a metric trade-off. The dense models exhibit a more mixed pattern. QMLP produces the shortest successful paths at 16.48\,m and improves over MLP in SPL and turn rate, but its lower success rate of 89\% suggests a less reliable policy that favors aggressive trajectory optimization at the expense of task completion.

These results highlight two main observations at the smallest environment scale. First, spiking-based agents already provide a clear advantage over the dense MLP-based models in terms of balanced navigation quality. Second, the proposed QSNN combines the reliability of the classical SNN with improved path efficiency and smoother motion, yielding the best overall trade-off among all evaluated methods in the 20$\times$20 setting.
\subsubsection{30$\times$30 Environment}

In the 30$\times$30 environment, all evaluated methods maintain high task-completion performance, with four out of five achieving at least 97\% success rate. However, the clearest separation between models appears in trajectory quality rather than in raw success alone. In this setting, the QSNN achieves the highest SPL at 0.917, the shortest mean path length at 26.13\,m, and the lowest turn rate at 0.251 turns per step, indicating the best balance between reliability, path efficiency, and motion smoothness. Although the Q-table baseline attains the highest success rate at 99\%, it produces the lowest SPL, the longest paths, and the highest turn rate, showing that frequent task completion does not necessarily translate into efficient or stable navigation.

A direct comparison between QSNN and its classical SNN counterpart again shows a clear advantage for the quantum-enhanced spiking architecture. Relative to SNN, QSNN improves SPL (0.917 vs.\ 0.887), path length (26.13\,m vs.\ 27.37\,m), and turn rate (0.251 vs.\ 0.292), with only a minor reduction in success rate (97\% vs.\ 98\%). This suggests that the quantum layer improves decision quality and motion regularity while preserving a highly competitive level of task completion. The dense models also benefit from quantum enhancement, although less uniformly. Compared with the classical MLP, QMLP improves success rate (98\% vs.\ 94\%), SPL (0.876 vs.\ 0.846), and turn rate (0.266 vs.\ 0.285), but it does not reduce path length, which remains slightly longer than that of the classical MLP.

These results indicate that, at the intermediate environment scale, success rate alone is not sufficient to identify the strongest navigation policy. While the Q-table baseline and QMLP remain highly competitive in raw completion, the QSNN provides the most favorable joint trade-off across all four metrics. This reinforces the pattern already observed in the 20$\times$20 setting, namely that the combination of spiking temporal processing and quantum feature transformation yields the most effective overall policy behavior.
\subsubsection{40$\times$40 Environment}

The 40$\times$40 environment is the most challenging setting and reveals the clearest separation between the evaluated agents. In this regime, the QSNN delivers the strongest overall performance, jointly achieving the highest success rate at 99\% and the highest SPL at 0.902. It also maintains a lower path length and turn rate than the other high-success agents, indicating that its policy remains both efficient and stable as the navigation task becomes more complex. By contrast, the classical MLP degrades sharply to a success rate of 77\%, showing that dense non-spiking architectures are substantially more affected by the increased environment scale and obstacle complexity.

A direct comparison between QSNN and its classical SNN counterpart again shows a consistent advantage for the quantum-enhanced spiking model. Relative to SNN, QSNN improves success rate (99\% vs.\ 98\%), SPL (0.902 vs.\ 0.885), path length (36.84\,m vs.\ 37.19\,m), and turn rate (0.260 vs.\ 0.324). These gains indicate that the quantum layer strengthens the spiking policy without introducing any trade-off across the four reported metrics. The gap between the dense models is even larger: QMLP substantially outperforms MLP in success rate (93\% vs.\ 77\%) and SPL (0.824 vs.\ 0.760), suggesting that quantum enhancement provides a marked reliability benefit in the dense setting as well.

Although the classical MLP attains the shortest path length and the lowest turn rate, these values should be interpreted cautiously. Since they are computed only over successful episodes, they reflect a strong survivorship effect: the model appears to succeed mainly on easier configurations while failing in a large fraction of more demanding cases. A similar point applies when comparing QSNN with the Q-table baseline. While both methods reach 99\% success rate, the Q-table policy yields substantially lower SPL, longer paths, and a much higher turn rate, indicating less efficient and more oscillatory behavior. The 40$\times$40 results reinforce the central trend of this study: as environment complexity increases, the QSNN preserves both reliability and trajectory quality more effectively than the competing approaches.
\subsection{Trajectory-Level Analysis}

The QSNN generates stable and goal-directed trajectories across all considered environment sizes. As the environment scale increases, the resulting paths become naturally longer and more structured, yet they remain smooth and continue to avoid both static and dynamic obstacles without visibly erratic motion. This qualitative behavior is consistent with the low turn-rate values and strong SPL scores reported in Table~\ref{tab:main_results}.

\begin{figure}[htpb]
    \centering
    \includegraphics[width=\linewidth]{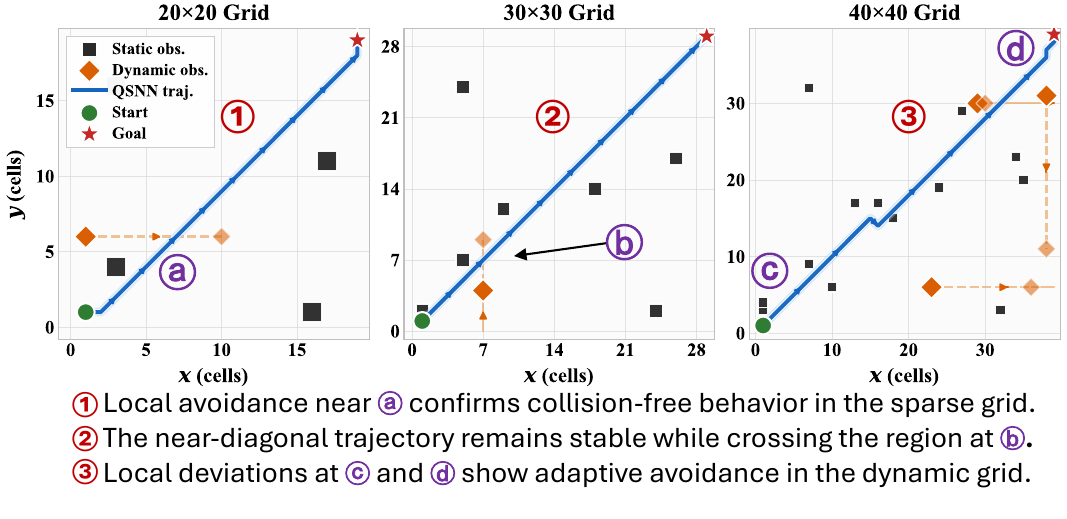}
\caption{Example QSNN trajectories in the 20$\times$20, 30$\times$30, and 40$\times$40 environments. Static obstacles are shown as dark cells, while dynamic obstacle motion is indicated by dashed traces.}
    \label{fig:qsnn_path_20x20}
\end{figure}

Figure~\ref{fig:qsnn_path_20x20} illustrates representative QSNN trajectories in the 20$\times$20, 30$\times$30, and 40$\times$40 environments. The figure shows that the learned policy maintains coherent obstacle-aware navigation across all tested scales, supporting the quantitative evidence that the QSNN preserves both efficiency and motion stability as task complexity increases.

\subsection{Quantum Hardware Evaluation}

To assess the feasibility of the proposed QSNN beyond simulation, we executed one navigation episode on IBM Quantum’s \texttt{ibm\_fez} backend using 1024 shots per circuit evaluation. This experiment is not intended as a statistically matched hardware benchmark, but rather as an initial verification that the learned hybrid policy can produce valid navigation behavior under real quantum execution conditions.
\begin{table}[htbp]
\centering
\caption{Inference metrics for one QSNN trajectory executed on IBM \texttt{ibm\_fez} quantum hardware.}
\label{tab:ibm_hardware}

\begin{tabular}{cccc}
\toprule
\rowcolor{headergray}
\textbf{Success Rate} & \textbf{SPL} & \textbf{Path Length (m)} & \textbf{Turn Rate} \\
\midrule
 1 & 0.8189 & 19.75 & 0.692 \\
\bottomrule
\end{tabular}
\end{table}
The hardware-executed QSNN policy successfully reached the target in this single episode, yielding a success value of 100\% for the tested run and an SPL of 0.8189, as reported in Table~\ref{tab:ibm_hardware}. The trajectory produced a path length of 19.75\,m and a turn rate of 0.692. Compared with the mean QSNN simulation results in the 20$\times$20 environment, the hardware run preserves task completion but shows reduced path efficiency and higher motion oscillation. Since the hardware result corresponds to only one episode, this comparison should be interpreted as an indicative feasibility check rather than a statistically equivalent evaluation. The observed gap is consistent with the expected effects of shot-based expectation estimation and hardware noise, which can perturb the action-value estimates used for greedy action selection.

\begin{figure}[htpb]
    \centering
    \includegraphics[width=1\linewidth]{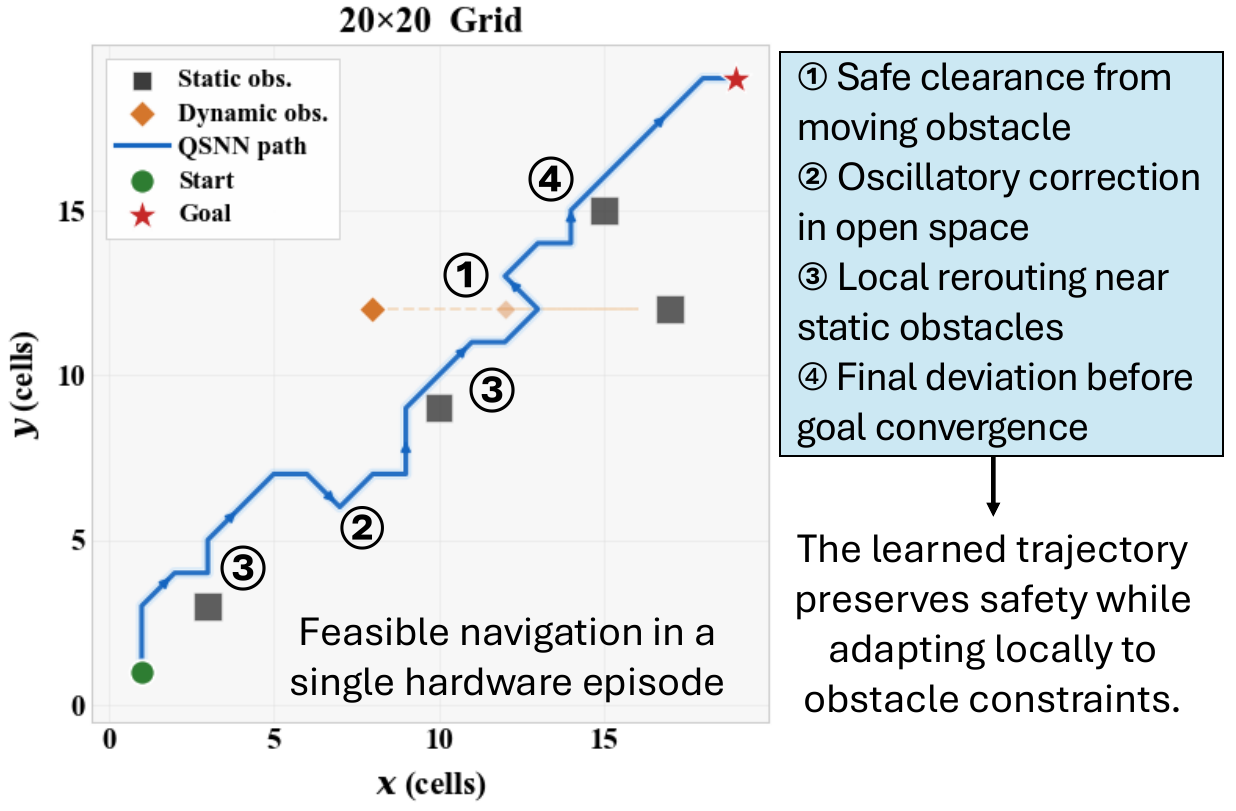}
\caption{Example QSNN trajectory executed on IBM \texttt{ibm\_fez} quantum hardware using 1024 shots per circuit evaluation.}
    \label{fig:ibm_traj}
\end{figure}

Figure~\ref{fig:ibm_traj} shows the trajectory obtained on hardware. Despite non-ideal execution conditions, the policy generates a valid obstacle-avoiding path and reaches the goal. These results provide an initial proof of feasibility for deploying the proposed quantum-enhanced spiking policy on real quantum hardware, while also highlighting the remaining gap between ideal simulation and current noisy quantum devices.

\subsection{Key Findings and Implications}

The main findings of this study can be summarized as follows:
\begin{itemize}
    \item Spiking architectures remain more robust than dense MLP-based policies across all evaluated environment sizes, with the performance gap becoming more pronounced as navigation complexity increases.
    
    \item The proposed QSNN provides the strongest overall trade-off between task completion, path efficiency, and motion smoothness, emerging as the best-performing model across the evaluated settings.
    
    \item The inclusion of a variational quantum layer is more effective in the spiking architecture than in the dense MLP-based one, suggesting that spike-based temporal processing offers a more suitable pre-quantum representation for this task.
    
    \item Although the Q-table baseline often achieves competitive success rates, it consistently produces longer and more oscillatory trajectories, showing that success rate alone is insufficient to characterize navigation quality.
    
    \item The successful execution of a QSNN trajectory on IBM quantum hardware supports the feasibility of the proposed hybrid approach, while also highlighting current limitations related to execution latency and shot-based variability.
\end{itemize}

\section{Conclusion}\label{sec5}

This work presented a fair and controlled study of classical, spiking, and quantum-enhanced reinforcement learning agents for grid-based navigation in environments containing both static and dynamic obstacles. The proposed framework combined a shared environment formulation, model-specific hyperparameter selection, deterministic Q-table conversion for unified inference, and evaluation across multiple environment scales using success rate, success-weighted path length, path length, and turn rate.

Across all evaluated settings, the QSNN achieved the strongest overall trade-off between task completion, trajectory efficiency, and motion smoothness, with its advantage becoming more evident as navigation complexity increased. The results further showed that spiking architectures remain more robust than dense MLP-based policies and that integrating a variational quantum layer into the spiking pipeline yields consistent gains over its classical counterpart. In addition, the successful execution of a QSNN trajectory on IBM quantum hardware provided an initial proof of feasibility for the proposed hybrid approach under real-device conditions.

Future work will extend this framework to richer navigation scenarios with more complex state representations, larger action spaces, and more realistic dynamic environments. Another important direction is the development of hardware-aware quantum circuit design and noise-resilient quantum-spiking integration strategies to further improve performance on near-term quantum processors.

\section*{Code Availability}
The code developed for this paper is publicly available at \url{https://github.com/eBrain4Everyone/Q-SpiRL}.

\section*{Acknowledgment}
 This work was supported in part by the NYUAD Center for Quantum and Topological Systems (CQTS), funded by Tamkeen under the NYUAD Research Institute grant CG008. This research was carried out on the High Performance Computing resources at New York University Abu Dhabi.
\bibliographystyle{IEEEtran}

\bibliography{refs}

@article{KUMAR2025129916,
  title={DSQN: Robust path planning of mobile robot based on deep spiking Q-network},
  author={Kumar, Aakash and Zhang, Lei and Bilal, Hazrat and Wang, Shifeng and Shaikh, Ali Muhammad and Bo, Lu and Rohra, Avinash and Khalid, Alisha},
  journal={Neurocomputing},
  volume={634},
  pages={129916},
  year={2025},
  publisher={Elsevier}
}

@inproceedings{11343694,
  title={The Emergence of Deep Reinforcement Learning for Path Planning},
  author={Nguyen, Thanh Thi and Nahavandi, Saeid and Razzak, Imran and Nguyen, Dung and Pham, Nhat Truong and Nguyen, Quoc Viet Hung},
  booktitle={2025 IEEE International Conference on Systems, Man, and Cybernetics (SMC)},
  pages={6265--6272},
  year={2025},
  organization={IEEE}
}

@article{tomar2025quantum,
  title={Quantum-Enhanced Hybrid Reinforcement Learning Framework for Dynamic Path Planning in Autonomous Systems},
  author={Tomar, Sahil and Alam, Shamshe and Kumar, Sandeep and Mathur, Amit},
  journal={arXiv preprint arXiv:2504.20660},
  year={2025}
}

@article{sinha2025nav,
  title={Nav-q: quantum deep reinforcement learning for collision-free navigation of self-driving cars},
  author={Sinha, Akash and Macaluso, Antonio and Klusch, Matthias},
  journal={Quantum Machine Intelligence},
  volume={7},
  number={1},
  pages={19},
  year={2025},
  publisher={Springer}
}

@article{hohenfeld2024quantum,
  title={Quantum deep reinforcement learning for robot navigation tasks},
  author={Hohenfeld, Hans and Heimann, Dirk and Wiebe, Felix and Kirchner, Frank},
  journal={IEEE Access},
  volume={12},
  pages={87217--87236},
  year={2024},
  publisher={IEEE}
}

@article{sarkar2024novel,
  title={A novel hybrid quantum architecture for path planning in quantum-enabled autonomous mobile robots},
  author={Sarkar, Mayukh and Pradhan, Jitesh and Singh, Anil Kumar and Nenavath, Hathiram},
  journal={IEEE Transactions on Consumer Electronics},
  volume={70},
  number={3},
  pages={5597--5606},
  year={2024},
  publisher={IEEE}
}

@article{yan2024quantum,
  title={Quantum robotics: a review of emerging trends},
  author={Yan, Fei and Iliyasu, Abdullah M and Li, Nianqiao and Salama, Ahmed S and Hirota, Kaoru},
  journal={Quantum Machine Intelligence},
  volume={6},
  number={2},
  pages={86},
  year={2024},
  publisher={Springer}
}

@article{sambataro2025current,
  title={Current Trends and advances in quantum navigation for maritime applications: A comprehensive review},
  author={Sambataro, Olga and Costanzi, Riccardo and Alves, Joao and Caiti, Andrea and Paglierani, Pietro and Petroccia, Roberto and Munaf{\`o}, Andrea},
  journal={IEEE Journal of Oceanic Engineering},
  year={2025},
  publisher={IEEE}
}

@article{dong2010robust,
  title={Robust quantum-inspired reinforcement learning for robot navigation},
  author={Dong, Daoyi and Chen, Chunlin and Chu, Jian and Tarn, Tzyh-Jong},
  journal={IEEE/ASME transactions on mechatronics},
  volume={17},
  number={1},
  pages={86--97},
  year={2010},
  publisher={IEEE}
}

@article{rathi2023exploring,
  title={Exploring neuromorphic computing based on spiking neural networks: Algorithms to hardware},
  author={Rathi, Nitin and Chakraborty, Indranil and Kosta, Adarsh and Sengupta, Abhronil and Ankit, Aayush and Panda, Priyadarshini and Roy, Kaushik},
  journal={ACM Computing Surveys},
  volume={55},
  number={12},
  pages={1--49},
  year={2023},
  publisher={ACM New York, NY}
}

@inproceedings{innan2025quav,
  title={{QUAV}: Quantum-Assisted Path Planning and Optimization for UAV Navigation with Obstacle Avoidance},
  author={Innan, Nouhaila and Kashif, Muhammad and Marchisio, Alberto and Gan, Yung-Sze and Barbaresco, Frederic and Shafique, Muhammad},
  booktitle={2025 IEEE International Conference on Quantum Artificial Intelligence (QAI)},
  pages={208--215},
  year={2025},
  organization={IEEE}
}

@inproceedings{chen2024qmarl,
  title={QMARL: A Quantum Multi-Agent Reinforcement Learning Framework for Swarm Robots Navigation},
  author={Chen, Weizhao and Wan, Jiawang and Ye, Fangwen and Wang, Ran and Xu, Cheng},
  booktitle={2024 IEEE International Conference on Acoustics, Speech, and Signal Processing Workshops (ICASSPW)},
  pages={388--392},
  year={2024},
  organization={IEEE}
}

@article{chen2025survey,
  title={A survey of autonomous robots and multi-robot navigation: Perception, planning and collaboration},
  author={Chen, Weinan and others},
    author2={Chen, Weinan and Chi, Wenzheng and Ji, Sehua and Ye, Hanjing and Liu, Jie and Jia, Yunjie and Yu, Jiajie and Cheng, Jiyu},
  journal={Biomimetic Intelligence and Robotics},
  volume={5},
  number={2},
  pages={100203},
  year={2025},
  publisher={Elsevier}
}

@article{nahavandi2025comprehensive,
  title={A comprehensive review on autonomous navigation},
  author={Nahavandi, Saeid and others},
    author2={Nahavandi, Saeid and Alizadehsani, Roohallah and Nahavandi, Darius and Mohamed, Shady and Mohajer, Navid and Rokonuzzaman, Mohammad and Hossain, Ibrahim},
  journal={ACM Computing Surveys},
  volume={57},
  number={9},
  pages={1--67},
  year={2025},
  publisher={ACM New York, NY}
}

@article{desouza2002vision,
  title={Vision for mobile robot navigation: A survey},
  author={DeSouza, Guilherme N and Kak, Avinash C},
  journal={IEEE transactions on pattern analysis and machine intelligence},
  volume={24},
  number={2},
  pages={237--267},
  year={2002},
  publisher={IEEE}
}

@book{sutton1998reinforcement,
  title={Reinforcement learning: An introduction},
  author={Sutton, Richard S and Barto, Andrew G and others},
  volume={1},
  number={1},
  year={1998},
  publisher={MIT press Cambridge}
}

@article{zhu2021deep,
  title={Deep reinforcement learning based mobile robot navigation: A review},
  author={Zhu, Kai and Zhang, Tao},
  journal={Tsinghua Science and Technology},
  volume={26},
  number={5},
  pages={674--691},
  year={2021},
  publisher={TUP}
}

@article{watkins1992q,
  title={Q-learning},
  author={Watkins, Christopher JCH and Dayan, Peter},
  journal={Machine learning},
  volume={8},
  number={3},
  pages={279--292},
  year={1992},
  publisher={Springer}
}

@article{mason2019review,
  title={A review of reinforcement learning for autonomous building energy management},
  author={Mason, Karl and Grijalva, Santiago},
  journal={Computers \& Electrical Engineering},
  volume={78},
  pages={300--312},
  year={2019},
  publisher={Elsevier}
}

@article{botvinick2019reinforcement,
  title={Reinforcement learning, fast and slow},
  author={Botvinick, Matthew and Ritter, Sam and Wang, Jane X and Kurth-Nelson, Zeb and Blundell, Charles and Hassabis, Demis},
  journal={Trends in cognitive sciences},
  volume={23},
  number={5},
  pages={408--422},
  year={2019},
  publisher={Elsevier}
}

@article{ghosh2009spiking,
  title={Spiking neural networks},
  author={Ghosh-Dastidar, Samanwoy and Adeli, Hojjat},
  journal={International journal of neural systems},
  volume={19},
  number={04},
  pages={295--308},
  year={2009},
  publisher={World Scientific}
}

@article{bing2018survey,
  title={A survey of robotics control based on learning-inspired spiking neural networks},
  author={Bing, Zhenshan and Meschede, Claus and R{\"o}hrbein, Florian and Huang, Kai and Knoll, Alois C},
  journal={Frontiers in neurorobotics},
  volume={12},
  pages={35},
  year={2018},
  publisher={Frontiers Media SA}
}

@article{zhang2025systematic,
  title={A systematic review of spiking neural networks for human-robot interaction in rehabilitative wearable robotics},
  author={Zhang, Xingyu and Cao, Yu and Huang, Jian and Liu, Jindong and Zhang, Zhi-Qiang},
  journal={IEEE Transactions on Cognitive and Developmental Systems},
  year={2025},
  publisher={IEEE}
}

@article{Biamonte2017,
  title={Quantum machine learning},
  author={Biamonte, Jacob and Wittek, Peter and Pancotti, Nicola and Rebentrost, Patrick and Wiebe, Nathan and Lloyd, Seth},
  journal={Nature},
  volume={549},
  number={7671},
  pages={195--202},
  year={2017},
  publisher={Nature Publishing Group UK London}
}

@article{chang2025primer,
  title={A primer on quantum machine learning},
  author={Chang, Su Yeon and Cerezo, M},
  journal={arXiv preprint arXiv:2511.15969},
  year={2025}
}

@article{Zanatta2024,
  title={Exploring spiking neural networks for deep reinforcement learning in robotic tasks},
  author={Zanatta, Luca and Barchi, Francesco and Manoni, Simone and Tolu, Silvia and Bartolini, Andrea and Acquaviva, Andrea},
  journal={Scientific Reports},
  volume={14},
  number={1},
  pages={30648},
  year={2024},
  publisher={Nature Publishing Group UK London}
}

@article{Chen2022DeepRL,
  title={Deep reinforcement learning with spiking q-learning},
  author={Chen, Ding and Peng, Peixi and Huang, Tiejun and Tian, Yonghong},
  journal={arXiv preprint arXiv:2201.09754},
  year={2022}
}

@article{Yang2025,
   author2 = {Bo Yang and Shibo Zhou and Chaohui Lin and Qingao Chai and Rui Yan and De Ma and Gang Pan and Huajin Tang},
    author = {Bo Yang and others},
   doi1 = {10.1109/ICRA55743.2025.11128063},
   isbn = {9798331541392},
   issn = {10504729},
   journal = {Proceedings - IEEE International Conference on Robotics and Automation},
   pages = {13384-13390},
   publisher = {Institute of Electrical and Electronics Engineers Inc.},
   title = {HSRL: A Hierarchical Control System Based on Spiking Deep Reinforcement Learning for Robot Navigation},
   url1 = {https://ieeexplore.ieee.org/document/11128063},
   year = {2025}
}

@article{innan2026spate,
  title={SPATE: Spiking-Phase Adaptive Temporal Encoding for Quantum Machine Learning},
  author={Innan, Nouhaila and Putra, Rachmad Vidya Wicaksana and Shafique, Muhammad},
  journal={arXiv preprint arXiv:2604.11022},
  year={2026}
}

@inproceedings{innan2025fl,
  title={{FL-QDSNNs}: Federated learning with quantum dynamic spiking neural networks},
  author={Innan, Nouhaila and Marchisio, Alberto and Shafique, Muhammad},
  booktitle={2025 IEEE International Conference on Quantum Artificial Intelligence (QAI)},
  pages={113--119},
  year={2025},
  organization={IEEE}
}

@inproceedings{Pasquali2023,
  title={Classification with Integrated Quantum and Spiking Neural Networks},
  author={Pasquali, Dominic and Grossi, Michele and Vallecorsa, Sofia},
  booktitle={2023 IEEE International Conference on Quantum Computing and Engineering (QCE)},
  volume={2},
  pages={298--299},
  year={2023},
  organization={IEEE}
}

@article{Khatoniar2024,
   author = {Ria Khatoniar and Debanjan Konar and Vaneet Aggarwal},
   doi1 = {10.1109/QCE60285.2024.10370},
   isbn = {9798331541378},
   journal = {Proceedings - IEEE Quantum Week 2024, QCE 2024},
   pages = {490-491},
   publisher = {Institute of Electrical and Electronics Engineers Inc.},
   title = {Quantum-Enhanced Spiking Neural Networks},
   volume = {2},
   url1 = {https://ieeexplore.ieee.org/document/10821021},
   year = {2024}
}

@inproceedings{Liu2025,
  title={Quantum spiking neural networks for image classification},
  author={Liu, Shilong and Gu, Yongjian},
  booktitle={Third International Conference on Algorithms, Network, and Communication Technology (ICANCT 2024)},
  volume={13545},
  pages={181--188},
  year={2025},
  organization={SPIE}
}

@article{Brand2024,
  title={A quantum leaky integrate-and-fire spiking neuron and network},
  author={Brand, Dean and Petruccione, Francesco},
  journal={npj Quantum Information},
  volume={10},
  number={1},
  pages={125},
  year={2024},
  publisher={Nature Publishing Group UK London}
}

@inproceedings{dutta2025qas,
  title={{QAS-QTNs}: Curriculum reinforcement learning-driven quantum architecture search for quantum tensor networks},
  author={Dutta, Siddhant and Innan, Nouhaila and Yahia, Sadok Ben and Shafique, Muhammad},
  booktitle={2025 IEEE International Conference on Quantum Computing and Engineering (QCE)},
  volume={1},
  pages={1739--1747},
  year={2025},
  organization={IEEE}
}

@inproceedings{dutta2024qadqn,
  title={Qadqn: Quantum attention deep q-network for financial market prediction},
  author={Dutta, Siddhant and Innan, Nouhaila and Marchisio, Alberto and Yahia, Sadok Ben and Shafique, Muhammad},
  booktitle={2024 IEEE International Conference on Quantum Computing and Engineering (QCE)},
  volume={2},
  pages={341--346},
  year={2024},
  organization={IEEE}
}

@article{JerbiQRL,
  title={Parametrized quantum policies for reinforcement learning},
  author={Jerbi, Sofiene and Gyurik, Casper and Marshall, Simon and Briegel, Hans and Dunjko, Vedran},
  journal={Advances in neural information processing systems},
  volume={34},
  pages={28362--28375},
  year={2021}
}

@article{meyer2022survey,
  title={A survey on quantum reinforcement learning},
  author={Meyer, Nico and Ufrecht, Christian and Periyasamy, Maniraman and Scherer, Daniel D and Plinge, Axel and Mutschler, Christopher},
  journal={arXiv preprint arXiv:2211.03464},
  year={2022}
}

@article{sefrin2025quantum,
  title={Quantum reinforcement learning in dynamic environments},
  author={Sefrin, Oliver and Radons, Manuel and Simon, Lars and W{\"o}lk, Sabine},
  journal={arXiv preprint arXiv:2507.01691},
  year={2025}
}

@inproceedings{innan2025lep,
  title={Lep-qnn: Loan eligibility prediction using quantum neural networks},
  author={Innan, Nouhaila and Marchisio, Alberto and Bennai, Mohamed and Shafique, Muhammad},
  booktitle={2025 IEEE International Conference on Quantum Computing and Engineering (QCE)},
  volume={1},
  pages={1864--1872},
  year={2025},
  organization={IEEE}
}

@article{innan2025qnn,
  title={QNN-VRCS: A Quantum Neural Network for Vehicle Road Cooperation Systems},
  author={Innan, Nouhaila and Behera, Bikash K and Al-Kuwari, Saif and Farouk, Ahmed},
  journal={IEEE Transactions on Intelligent Transportation Systems},
  year={2025},
  publisher={IEEE}
}

@inproceedings{innan2025next,
  title={Next-Generation Quantum Neural Networks: Enhancing Efficiency, Security, and Privacy},
  author={Innan, Nouhaila and Kashif, Muhammad and Marchisio, Alberto and Bennai, Mohamed and Shafique, Muhammad},
  booktitle={2025 IEEE 31st International Symposium on On-Line Testing and Robust System Design (IOLTS)},
  pages={1--4},
  year={2025},
  organization={IEEE}
}

\end{document}